\RequirePackage{fix-cm}
\documentclass[a4paper,11pt]{article}

\usepackage{authblk}
\usepackage{algorithm}
\usepackage{algorithmic}
\usepackage{multirow}
\usepackage{graphicx}
\usepackage{booktabs}
\usepackage{epsfig}
\usepackage{epstopdf}
\usepackage{amsmath}
\usepackage{caption}
\usepackage{subcaption}
\usepackage{amsfonts}
\usepackage{wasysym}
\usepackage{paralist}
\usepackage{url}
\usepackage{float}
\usepackage{booktabs}
\usepackage{microtype}
\usepackage{paralist}
\usepackage{pifont}
\newcommand{\cmark}{\ding{51}}%
\newcommand{\xmark}{\ding{55}}%

\usepackage{etoolbox}

\providecommand{\keywords}[1]{\textbf{\textit{Keywords }} #1}

\begin{document} 

\title{Digging Deeper: Operator Analysis for Optimizing Nonlinearity of Boolean Functions}

\author[1]{Marko Djurasevic}
\author[1]{Domagoj Jakobovic}
\author[2]{Luca Mariot}
\author[3]{Stjepan Picek}

\affil[1]{{\small Faculty of Electrical Engineering and Computing, University of Zagreb, Unska 3, Zagreb, Croatia}
	
	{\small \texttt{\{marko.durasevic,domagoj.jakobovic\}@fer.hr}}}

\affil[2]{{\small Semantics, Cybersecurity and Services Group, University of Twente, Drienerlolaan 5, 7522 NB Enschede, The Netherlands}
	
	{\small \texttt{l.mariot@utwente.nl}}}

\affil[3]{{\small Digital Security Group, Radboud University, PO Box 9010, 6500 GL Nijmegen, The Netherlands} 
	
	{\small \texttt{stjepan.picek@ru.nl}}}

\maketitle

\begin{abstract}
Boolean functions are mathematical objects with numerous applications in domains like coding theory, cryptography, and telecommunications. Finding Boolean functions with specific properties is a complex combinatorial optimization problem where the search space grows super-exponentially with the number of input variables. One common property of interest is the nonlinearity of Boolean functions. Constructing highly nonlinear Boolean functions is difficult as it is not always known what nonlinearity values can be reached in practice. In this paper, we investigate the effects of the genetic operators for bit-string encoding in optimizing nonlinearity. While several mutation and crossover operators have commonly been used, the link between the genotype they operate on and the resulting phenotype changes is mostly obscure. By observing the range of possible changes an operator can provide, as well as relative probabilities of specific transitions in the objective space, one can use this information to design a more effective combination of genetic operators.
The analysis reveals interesting insights into operator effectiveness and indicates how algorithm design may improve convergence compared to an operator-agnostic genetic algorithm.
\end{abstract}


\section{Introduction}
\label{sec:introduction}

Boolean functions are used in many applications, including combinatorial designs~\cite{Rothaus}, coding theory~\cite{KERDOCK1972182}, sequences~\cite{1056589}, telecommunications~\cite{Paterson}, and cryptography~\cite{https://doi.org/10.48550/arxiv.2301.08012}. For Boolean functions to be useful, they need to fulfill certain properties. One of the core properties considered across multiple domains is nonlinearity, and the requirement is to have it as high as possible.
To design such functions, one can employ algebraic constructions, random search, and metaheuristics.
With metaheuristics, it is important to consider what will be the solution encoding and objective function. The most common choice to encode Boolean functions is to use the truth table encoding. Unfortunately, the problem is that by considering the truth table, it is difficult to assess the quality of most of the relevant properties, including nonlinearity. To evaluate the nonlinearity, one first needs to translate the truth table into the Walsh-Hadamard spectrum and then calculate the nonlinearity value.

Metaheuristics commonly operate by constructing solutions in one way but evaluating their quality in another way, where the connection is not necessarily straightforward.
The problem becomes even more challenging when considering larger Boolean functions as the search space size equals $2^{2^n}$ for a Boolean function with $n$ inputs. Thus, even small Boolean functions with, e.g., 7 inputs would result in a huge search space of $2^{128}$. Then, the argument that metaheuristics does not work well because the considered Boolean functions are too large sounds natural. Moreover, certain properties can be computationally expensive to evaluate\footnote{For instance, the naive implementation of the Walsh-Hadamard transform has complexity $\mathcal O(2^{2n})$ while the optimized butterfly algorithm has complexity $\mathcal O(n2^n)$.}  for larger Boolean sizes, making it difficult to be checked often in the optimization process. Alleviating any of those issues is important as it could allow finding more well-performing Boolean functions in different sizes, having practical ramifications for diverse application domains.

While the aforesaid reasons are natural difficulties when using metaheuristics, potential solutions are different. For instance, considering the computational complexity of evaluations, a solution can be to ``simply'' parallelize the objective function calculation and/or the search algorithm~\cite{10.1007/978-3-319-10762-2_41}. To facilitate optimization of Boolean functions of larger size, besides more efficient calculations, another option could be to design constructions of Boolean functions~\cite{10.1145/2908812.2908915} that work for any Boolean function size. 
Finally, to avoid the problems between the genotype and phenotype mapping/evaluation, one could consider the intuitive option of working with the ``correct'' encoding from the beginning. There are efforts like that, see, e.g.,~\cite{1299941}, but unfortunately, there are still issues since working with the Walsh-Hadamard encoding of solutions will commonly result only in pseudo-Boolean functions, limiting the practicality of the approach.
Interestingly, we found no works that aim to understand what is the influence of a certain change (e.g., stemming from crossover or mutation) in genotype (truth table) to phenotype (Walsh-Hadamard spectrum).

This work aims to fill in this gap and assess the influence that some common crossover and mutation operators have on the nonlinearity property changes in the genetic algorithm (GA) and local search (LS). We consider two objective functions, eight mutation operators, and two crossover operators. We conduct our analysis by exhaustively checking the space of Boolean functions with three and four inputs and by sampling for larger Boolean function sizes.
Our main contributions are:
\begin{compactenum}
\item We are the first to systematically evaluate different mutation and crossover operators for evolving Boolean functions with high nonlinearity.
\item We show how simple mutation operators can improve the nonlinearity value and how a more informative fitness function helps where operators do not work well.
\item We observe that highly successful crossovers occur only for parents of lower nonlinearity and when both parents have a similar nonlinearity.
\item  We demonstrate that LS can outperform GA, and it benefits more from the additional information in the fitness function.
\end{compactenum}

\section{Preliminaries on Boolean Functions}
\label{sec:boolean}

Let $n$ be a positive integer, i.e.,  $n \in \mathbb{N}^+$.
We denote by $\mathbb{F}_{2}^{n}$ the $n$-dimensional vector space over $\mathbb{F}_{2}$ and by $\mathbb{F}_{2^n}$ the finite field with $2^n$ elements (i.e., of order $2^n$).
The set of all $n$-tuples of elements in the field $\mathbb{F}_{2}$ is denoted by $\mathbb{F}_{2}^{n}$, where $\mathbb{F}_{2}$ is the Galois field with two elements. 
The usual inner product of $a$ and $b$ equals $a\cdot b = \bigoplus_{i=1}^{n} a_{i}b_{i}$ in $\mathbb{F}_2^n$.
Since for every $n$, there exists a field $\mathbb{F}_{2^n}$ of order $2^n$, which is an $n$-dimensional vector space, we can endow the vector space $\mathbb{F}_2^n$ with the structure of that field when convenient.
Adding elements of the finite field $\mathbb{F}_{2^n}$ is denoted ``+'', as usual in mathematics. Since often, we identify $\mathbb{F}_{2}^n$ with $\mathbb{F}_{2^n}$, and if there is no ambiguity, we denote the addition of vectors of $\mathbb{F}_{2}^n$ when $n>1$ ``+'' as well.

An $n$-variable Boolean function is a mapping $f: \mathbb F_2^n \to \mathbb F_2$.
A Boolean function $f$ can be uniquely represented by the truth table that is the list of pairs of function inputs $x \in \mathbb F_2^n$ and corresponding function values $f(x)$. 
The value vector is the binary vector $\Omega_f$ composed of all $f(x)$, with $ x \in \mathbb{F}_2^n$, where some total order has been fixed on $\mathbb{F}_2^n$ (commonly, the lexicographic order). The size of the value vector is $2^n$, and the size of the search space equals $2^{2^n}$.

A second option to uniquely represent a Boolean function $f$ is the Walsh-Hadamard Transform:
\begin{equation}
\label{eq:wht}
W_{f} (a) = \sum\limits_{x \in \mathbb F_{2}^{n}} (-1)^{f(x) \oplus a\cdot x}.
\end{equation}
The coefficient $W_f(a)$ measures the correlation between $f$ and the linear function $a \cdot x$. 
Notice that the mapping $W_f$ is injective, from which it follows that the spectrum of a Boolean function $f$ uniquely identifies $f$. In particular, one can retrieve the truth table of $f$ from its Walsh-Hadamard representation by using the Inverse Walsh-Hadamard Transform.\footnote{Inverse Walsh-Hadamard Transform follows Eq.~\eqref{eq:wht}, except that $x$ and $(-1)^{f(x)}$ are replaced by $a$ and $W_f(a)$, respectively, and the sum is normalized by a $2^{-n}$ factor.}

The minimum Hamming distance\footnote{The Hamming distance between two functions $f$ and $g$ is the size of the set $\left\lbrace x \in \mathbb{F}_2^n: f(x) \neq g(x) \right\rbrace$.} between a Boolean function $f$ and all affine functions is called the nonlinearity of $f$, denoted by $nl_{f}$. This property can be characterized in terms of the Walsh-Hadamard coefficients as follows:
\begin{equation}
\label{eq:nonlinearity}
nl_{f} = 2^{n - 1} - \frac{1}{2}\max_{a \in \mathbb F_{2}^{n}} |W_{f}(a)|.
\end{equation}

A common requirement is that nonlinearity should be as large as possible. From Eq.~\eqref{eq:nonlinearity}, this happens if the largest absolute value Walsh-Hadamard coefficient is as small as possible. The Parseval's relation $\sum_{a\in {\mathbb F}_2^n}W_f(a)^2=2^{2n}$ implies that the mean of $W_f(a)^2$ equals $2^n$. Finally, $\max_{a\in {\mathbb F}_2^n} |W_f(a)|$ is then at least equal to the square root of this mean. This allows for deriving the following inequality, known as the covering radius bound:
\begin{equation}
\label{eq_boolean_covering}
    nl_{f} \leq 2^{n-1}-2^{n/2-1}.
\end{equation}

A Boolean function can be considered highly nonlinear if its nonlinearity is close to the covering radius bound. The functions whose nonlinearity equals the maximal value $2^{n-1}-2^{n/2-1}$ are called bent. Bent functions exist only for even values of $n$. 

\section{Related Work}
\label{sec:related}

A literature review reveals that nonlinearity is the most explored property when optimizing Boolean functions.
Most of the works using metaheuristics to optimize properties do not try to investigate the influence of recombination operators but concentrate on finding Boolean functions in specific dimensions and property values. For a more recent overview of metaheuristic techniques for constructing Boolean functions, we refer readers to~\cite{https://doi.org/10.48550/arxiv.2301.08012}.

Millan et al. were the first to apply genetic algorithms (GAs) to evolve Boolean functions with good cryptographic properties~\cite{Millan97}. The authors used a genetic algorithm to evolve Boolean functions with high nonlinearity.
Dawson et al. used a combination of simulated annealing and hill-climbing with a cost function motivated by the Parseval theorem to find functions with high nonlinearity and low
autocorrelation~\cite{clark_two_stage}.
Aguirre et al. were the first to use a multi-objective random bit climber to search for balanced (having the same number of zeros and ones in the truth table) Boolean functions with high nonlinearity~\cite{hernan}.
Picek et al. were the first to use genetic programming to find Boolean functions with high nonlinearity (alongside more properties)~\cite{picekgecco2013}. 
Mariot and Leporati proposed using Particle Swarm Optimization to find Boolean functions with good trade-offs of cryptographic properties~\cite{MariotL15}. Manzoni et al. systematically investigated crossover operators that preserve balancedness and applied them to maximize the nonlinearity of Boolean functions~\cite{manzoni20}.
These works either use the truth table encoding or symbolic encoding mapped to the truth table. 

Picek and Jakobovic were the first to propose using evolutionary algorithms (genetic programming) to evolve algebraic constructions of Boolean functions~\cite{10.1145/2908812.2908915}. Carlet et al. followed a similar principle but concentrated on algebraic constructions fulfilling more than one property~\cite{10.1145/3512290.3528871}. While such works do not use the truth table encoding in the design of solutions, once the construction is evolved, it is applied to get a truth table of a Boolean function, which is then evaluated in the same way as in the previously discussed works.

Stepping away from the truth table encoding (or symbolic one), several works tried to evolve Boolean functions with good properties by using the Walsh-Hadamard spectrum as the solution encoding. While an intuitive approach, the issue here is that the Walsh-Hadamard transform is injective, so using a random Walsh-Hadamard spectrum will most likely result in pseudo-Boolean functions. 
The first work that considered using the Walsh-Hadamard spectrum to encode solutions is by Clark et al.~\cite{1299941}. While less ``popular'' than truth table encoding discussed before, several more works appeared that give good results but generally cannot compete with the truth table encoding~\cite{StanicaMC04, 10.1007/978-3-319-26841-5_3}.

To the best of our knowledge, there are only a few papers concentrating on the difficulties when using metaheuristics to construct Boolean functions with good properties like nonlinearity.
Picek et al. investigated the symmetry structure and fitness landscapes in the bit-string representation (truth table encoding) of Boolean functions~\cite{picekgeccop12015}. The authors concluded that due to a large number of symmetries, crossover behaves as a macro-mutation until the symmetries have been broken.
Picek et al. conducted fitness landscape and deception analysis and found no significant differences that could justify the difficulty in increasing the Boolean function size from 6 to 8 inputs~\cite{7744197}. 
Jakobovic et al. used fitness landscape analysis based on Local Optima Networks (LONs) and investigated the influence of several optimization criteria and variation operators~\cite{JAKOBOVIC2021107327}. The authors considered the truth table encoding and three neighborhood variants.

\section{Operator Analysis}
\label{sec:analysis}

The analysis here is performed exhaustively for functions with 3 and 4 variables and by sampling for larger sizes.
All the operators are defined for bit-string representation, with Boolean functions encoded in the truth table form.

\subsection{Problem Statement}
\label{sec:problem}

When optimizing Boolean functions to have a high nonlinearity, we require the Walsh-Hadamard spectrum to calculate nonlinearity. In this context, the spectrum corresponds to the phenotype of candidate solutions. 
However, if the solution encoding is different from the Walsh-Hadamard, it becomes unclear how the change in the genotype maps to the phenotype. As most works use the truth table encoding for metaheuristics, there is a problem with properly assessing the influence of a certain change (i.e., the result of a variation operator) in the solution.
This motivates the question of how a specific variation operator change in the genotype changes the phenotype and whether we can find some changes that can be observed for both genotype and phenotype. Indeed, consider a case where a mutation operator flips a single bit in the truth table encoding. Consequently, the Walsh-Hadamard spectrum will change, but the change will not necessarily align concerning the magnitude or position.

\subsection{Fitness Functions for Optimization of Boolean Functions}

Several objective functions can be defined to optimize Boolean function nonlinearity regardless of how the optimization is performed. These fitness functions were selected based on the literature study of common choices in related works~\cite{https://doi.org/10.48550/arxiv.2301.08012}. More options are possible, but they commonly include additional weight factors, making the operator analysis more complex.
The first fitness function is the simplest one and maximizes the nonlinearity value:
\begin{equation}
\label{eq:first}
    fitness_1 : nl_f.
\end{equation}

The second fitness function extends the first one to consider the whole Walsh-Hadamard spectrum and not only its extreme value (see Eq.~\eqref{eq:nonlinearity}).
Here, we count the number of occurrences of the maximal absolute value in the spectrum, denoted as $\#max\_values$.
As higher nonlinearity corresponds to a lower maximal absolute value, we aim for as few occurrences of the maximal value as possible in the hope it would be easier for the algorithm to reach the next nonlinearity value.
In this way, we provide the algorithm with additional information, making the objective space more gradual.
With this in mind, the second fitness function is defined as:

\begin{equation}
\label{eq:second}
fitness_2 : nl_{f} + \frac{2^n - \#max\_values}{2^n}.
\end{equation}
The second term never reaches the value of $1$ since, in that case, we effectively reach the next nonlinearity level.

\subsection{Mutation Operators}

We experimented with the mutation operators listed in Table~\ref{tab:operators}. The table also reports the neighborhood sizes of these operators if used in the local search.
All these operators are commonly used in evolutionary algorithms, although with different levels of efficiency.

\begin{table}
\caption{Selected mutation operators and their neighborhood sizes}
\label{tab:operators}
\centering
\begin{tabular}{@{}lll@{}}
\toprule
Mutation operator                & Neighborhood size &  \\ \midrule
bit set (0 to 1)                 & $\le n$           &  \\
bit reset (1 to 0)               & $\le n$           &  \\
bit flip (1 bit inverted)        & $n$               &  \\
two bit flip                     & ${n(n-1)}/{2}$                  &  \\
two bit flip, only if both equal & $\le {n(n-1)}/{2}$                  &  \\
two bit set                      & $\le {n(n-1)}/{2}$                  &  \\
two bit reset                    & $\le {n(n-1)}/{2}$                  &  \\
rotation                         & $n-1$               &  \\ \bottomrule
\end{tabular}
\end{table}

\subsection{Changes in the Walsh-Hadamard Spectrum}

As the first question in the analysis, we investigate whether a specific genotype (truth table) modification will yield a \textit{consistent} change in the phenotype (the Walsh-Hadamard spectrum, denoted for brevity spectrum), which can be utilized to maximize the objective value.
To answer this, we perform an exhaustive search for all mutation operators. For each operator, a mutation with every possible \textit{position} is performed on every possible Boolean function.
Here, a mutation position may refer to a single bit, two bit positions, or a number of rotations, depending on the operator.
Then, after each genotype change, we record new values of the spectrum and check whether the \textit{differences} in values of the spectrum were the same for the same mutation position.
For example, if inverting a single bit at position 0 (the least significant bit in our implementation) yields the same changes in every element of the spectrum, this is considered a \textit{consistent} phenotype change.
As it turns out, this phenomenon is only evident with simple operators that either only set or reset certain bits: bit set/reset and two bit set/reset since they result in a consistent change in spectrum values for any Boolean function.
For example, for $n=3$ and bit set, we observe the changes in the spectrum presented in Table~\ref{tab:WHbitset3}.

\begin{table}
\small
\caption{Changes in the Walsh-Hadamard spectrum: bit set operator, $n=3$}
\label{tab:WHbitset3}
\centering
\begin{tabular}{l rrrrrrrr }
\toprule
bit set &   \multicolumn{8}{c}{Changes in spectrum values}     \\ \midrule
0   & -2 & -2 & -2 & -2 & -2 & -2 & -2 & -2  \\
1   & -2 & 2  & -2 & 2  & -2 & 2  & -2 & 2   \\
2   & -2 & -2 & 2  & 2  & -2 & -2 & 2  & 2   \\
3   & -2 & 2  & 2  & -2 & -2 & 2  & 2  & -2  \\
4   & -2 & -2 & -2 & -2 & 2  & 2  & 2  & 2   \\
5   & -2 & 2  & -2 & 2  & 2  & -2 & 2  & -2  \\
6   & -2 & -2 & 2  & 2  & 2  & 2  & -2 & -2  \\
7   & -2 & 2  & 2  & -2 & 2  & -2 & -2 & 2  \\ \bottomrule
\end{tabular}
\end{table}

The reset operator results in the changes with the same values but with the opposite sign; the same is true for two bit set and reset (we omit presenting their patterns as they depend on two bit positions, which yields $n(n-1)/2$ combinations).
Depending on the initial spectrum, a single bit set/reset will either increase or decrease the nonlinearity by one, which over the set of all possible functions occurs in the equal number of cases, while the two bit operators modify nonlinearity by $\pm 2$.
The above patterns are present in a higher number of variables and can be generalized to any Boolean function size.

For some Boolean functions, it is possible to use this information in the following way: given the initial spectrum, we can choose a specific change pattern that would make the transition to a spectrum with a smaller maximal absolute value and consequently with a higher nonlinearity.
Unfortunately, this is \textit{not} possible for every function; one can easily verify that in some cases, as discussed in the next section (when nonlinearity is not already at the maximum value), no change using these operators can make the transition to a function with a higher nonlinearity.

\subsection{Probability of Changes in Nonlinearity}

Since a consistent change is evident only for a few operators, and by itself, their neighborhood cannot always be used to traverse to a higher nonlinearity, we tried to answer the following question: what is the probability that a specific mutation will result in a higher (or same or lower) nonlinearity?
In this case, we again performed an exhaustive search where, for a specific operator, we iterated over every mutation position and every Boolean function of the \textit{same} given nonlinearity.
For each change, we record the outcome and accumulate over all functions with the same nonlinearity.

In this experiment, we restricted the analysis to three operators: single bit flip, two bit flip, and rotation.
The reason for this is that the solutions from those operators include the solutions for all the remaining mutation operators in their neighborhood.
The results for bit flip and two bit flip reveal an interesting characteristic; over all functions with the same initial nonlinearity, the cumulative probabilities of nonlinearity changes are the same \textit{regardless} of the mutation position!
For instance, for $n=4$ and over all the functions with initial nonlinearity of 2, the bit flip operator on \textit{any} position will have the same probability of increasing the nonlinearity; this holds for every initial nonlinearity value.
The resulting probabilities for bit flip and two bit flip are shown in Table~\ref{tab:prob_n4}.

\begin{table}
\small
\centering
\caption{Probabilities of change (increase/no change/decrease) in nonlinearity for different starting value and mutation operator, $n=4$}
\begin{tabular}{@{}lllllll@{}}
\toprule
\multicolumn{7}{c}{starting $nl_f$} \\ 
0       & 1       & 2       & 3       & 4       & 5       & 6       \\ \midrule
\multicolumn{7}{c}{bit flip}                                                       \\
             100/0/0 & 93/0/7  & 87/0/13 & 81/0/19 & 48/0/52 & 7/0/93  & 0/0/100 \\
\multicolumn{7}{c}{two bit flip}                                                   \\
             100/0/0 & 87/13/0 & 75/25/0 & 40/57/3 & 4/86/10 & 0/50/50 & 0/0/100 \\ \bottomrule
\end{tabular}

\label{tab:prob_n4}
\end{table}

\begin{figure}
    \centering
    \includegraphics[width=0.6\textwidth]{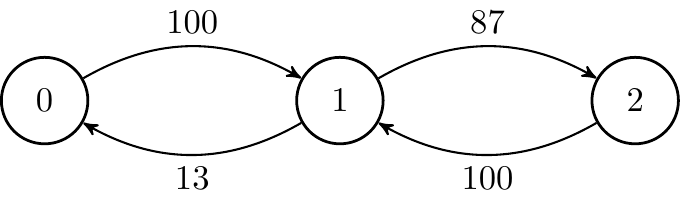}
    \caption{Nonlinearity transitions for bit flip, $n=3$}
    \label{fig:markov_bitflip_3}
\end{figure}

\begin{figure}
    \centering
    \includegraphics[width=0.6\textwidth]{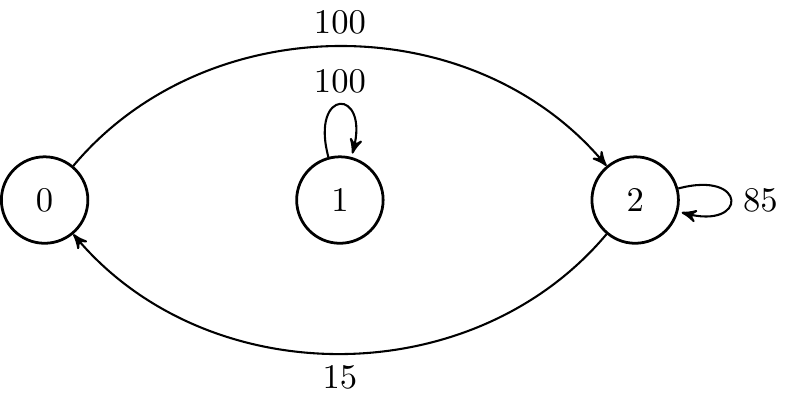}
    \caption{Nonlinearity transitions for two bit flip, $n=3$}
    \label{fig:markov_2bitflip_3}
\end{figure}

\begin{figure*}
    \centering
    \includegraphics[width=\textwidth]{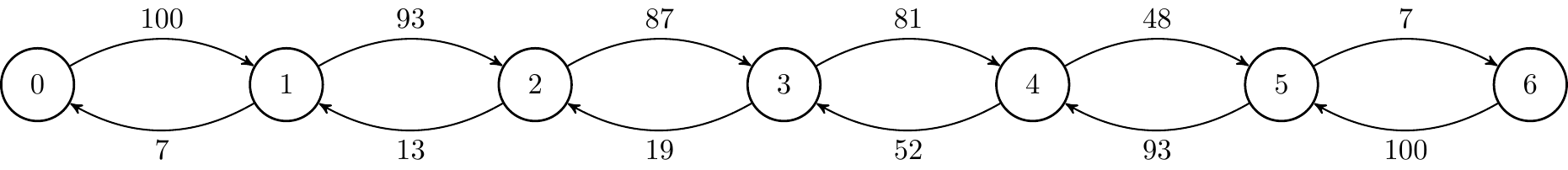}
    \caption{Nonlinearity transitions for bit flip, $n=4$}
    \label{fig:markov_bitflip_4}
\end{figure*}

\begin{figure*}
    \centering
    \includegraphics[width=\textwidth]{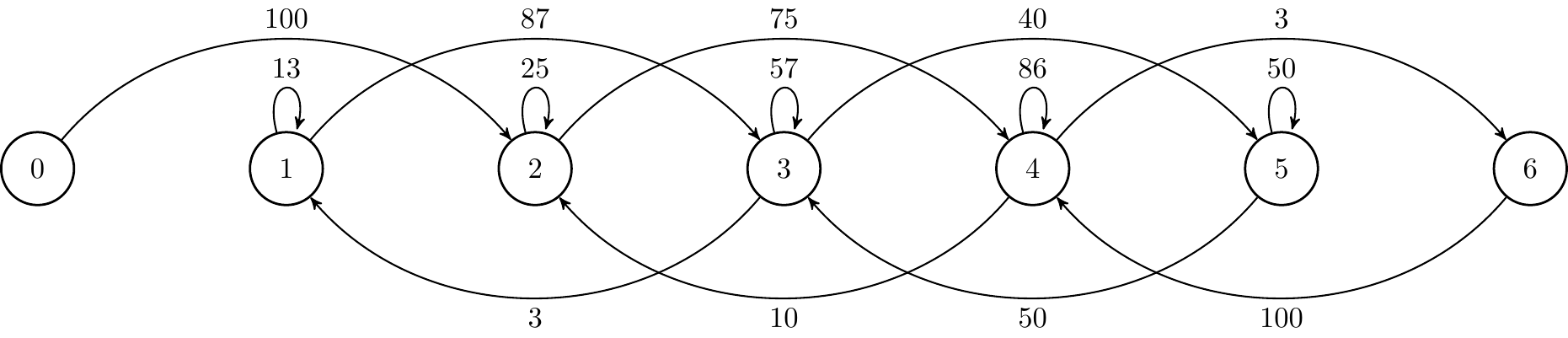}
    \caption{Nonlinearity transitions for two bit flip, $n=4$}
    \label{fig:markov_2bitflip_4}
\end{figure*}

The values in the table are shown in percentages so that the numbers in the triple respectively indicate the probability of increase, no change, and decrease in nonlinearity. 
The mutation bit position is omitted because the probabilities add to the same amounts for all mutation positions.
This behavior allows us to depict it in the form of a Markov chain; the transitions for all the possible nonlinearity values in Boolean functions of $n=3$ variables are shown in Figures~\ref{fig:markov_bitflip_3} and~\ref{fig:markov_2bitflip_3} and for $n=4$ in Figures~\ref{fig:markov_bitflip_4} and~\ref{fig:markov_2bitflip_4}.

In contrast to this, the probabilities for the rotation operator depend on the number of rotated bits; this operator seems to adhere to the following pattern:
\begin{compactitem}
    \item the rotation makes no change in nonlinearity for all shifts in $n \cdot i$ bits;
    \item for other cases, the probabilities are the same for all shifts in odd and even numbers of bits.
\end{compactitem}
For example, for $n=4$ the rotation produces changes with the probabilities shown in Table~\ref{tab:markov_rot_4}
(the first column represents the number of rotated bits).

\begin{table}
\caption{Probabilities of change (increase/no change/decrease) in nonlinearity for rotation operator, $n=4$}
\label{tab:markov_rot_4}
\centering
{\small
\begin{tabular}{@{}llllllll@{}}
\toprule
rot    & \multicolumn{7}{c}{starting $nl_f$}                                     \\ 
    \cmidrule{1-1} \cmidrule(lr){2-8}
 & 0       & 1       & 2       & 3       & 4       & 5       & 6       \\
1   & 75/25/0 & 68/31/0 & 57/41/0 & 31/66/1 & 2/89/7  & 0/60/39 & 0/21/78 \\
2   & 50/50/0 & 50/50/0 & 40/60/0 & 22/75/1 & 1/92/5  & 0/71/28 & 0/42/57 \\
3   & 75/25/0 & 68/31/0 & 57/41/0 & 31/66/1 & 2/89/7  & 0/60/39 & 0/21/78 \\
4   & 0/100/0 & 0/100/0 & 0/100/0 & 0/100/0 & 0/100/0 & 0/100/0 & 0/100/0 \\
5   & 75/25/0 & 68/31/0 & 57/41/0 & 31/66/1 & 2/89/7  & 0/60/39 & 0/21/78 \\
6   & 50/50/0 & 50/50/0 & 40/60/0 & 22/75/1 & 1/92/5  & 0/71/28 & 0/42/57 \\
7   & 75/25/0 & 68/31/0 & 57/41/0 & 31/66/1 & 2/89/7  & 0/60/39 & 0/21/78 \\
8   & 0/100/0 & 0/100/0 & 0/100/0 & 0/100/0 & 0/100/0 & 0/100/0 & 0/100/0 \\
9   & 75/25/0 & 68/31/0 & 57/41/0 & 31/66/1 & 2/89/7  & 0/60/39 & 0/21/78 \\
10  & 50/50/0 & 50/50/0 & 40/60/0 & 22/75/1 & 1/92/5  & 0/71/28 & 0/42/57 \\
11  & 75/25/0 & 68/31/0 & 57/41/0 & 31/66/1 & 2/89/7  & 0/60/39 & 0/21/78 \\
12  & 0/100/0 & 0/100/0 & 0/100/0 & 0/100/0 & 0/100/0 & 0/100/0 & 0/100/0 \\
13  & 75/25/0 & 68/31/0 & 57/41/0 & 31/66/1 & 2/89/7  & 0/60/39 & 0/21/78 \\
14  & 50/50/0 & 50/50/0 & 40/60/0 & 22/75/1 & 1/92/5  & 0/71/28 & 0/42/57 \\
15  & 75/25/0 & 68/31/0 & 57/41/0 & 31/66/1 & 2/89/7  & 0/60/39 & 0/21/78 \\ \bottomrule
\end{tabular}
}
\end{table}

In principle, the above information could allow the algorithm, given the current nonlinearity, to select the operator that will result in a higher nonlinearity with the highest possible probability.
Unfortunately, while this may increase the efficiency, it does not guarantee obtaining a higher nonlinearity value for all starting Boolean functions, as discussed next.

\subsection{Effect of Neighborhoods}

The fact that it is possible to traverse from any $nl_f$ to a higher nonlinearity does not mean this is possible for \textit{all} functions with a given $nl_f$. 
To see the limitations of these operators, we performed the following: for a given value of $nl_f$, we iterated over every function with this nonlinearity and over every operator and every mutation position.
For each function and operator pair, we only recorded whether it is possible to reach (any) higher nonlinearity using that operator.
Since there are three operators (bit flip, two bit flip, and rotation), this gives eight possibilities regarding each operator's application success for a given function.

For Boolean functions with three variables, for every function of every nonlinearity, there is at least one of these operators that can be applied to increase nonlinearity.
However, this is not true for $n=4$, for which the results are presented in Table~\ref{tab:succ_n4}.
Each row in the table reports the number of functions (with $nl_f$ in that column) for which it was \textit{possible or not} to get higher $nl_f$ with all operators in that row.
Ignoring the nonlinearity of 6, which is the highest possible value, there are 1120 functions of nonlinearity 4, for which neither operator can transition to a higher nonlinearity function.
These functions represent the dead-end of any local search with these neighborhoods (later verified in the experiments, as local search converges to the nonlinearity of either 4 or 6).

\begin{table}
\small
\centering
\caption{Number of functions with starting $nl_f$ where selected mutation operators can increase nonlinearity, $n=4$}
\label{tab:succ_n4}
\begin{tabular}{@{}ccccccccc@{}}
\toprule
\multicolumn{3}{c}{Operator}               & \multicolumn{6}{c}{starting $nl_f$}             \\ \cmidrule(lr){1-3} \cmidrule(lr){4-9}
rot        & bit flip       & 2 bit flip      & 0  & 1   & 2    & 3     & 4     & 5        \\ \midrule
\cmark        & \cmark           & \cmark           & 24 & 384 & 2688 & 8704  & 1408  & 0        \\
\cmark        & \cmark           & \xmark            & 0  & 0   & 0    & 0     & 0     & 0        \\
\cmark        & \xmark & \cmark           & 0  & 0   & 0    & 0     & 0     & 0        \\
\cmark        & \xmark            & \xmark            & 0  & 0   & 0    & 0     & 0     & 0        \\
\xmark         & \cmark           & \cmark           & 8  & 128 & 1152 & 9216  & 25472 & 0        \\
\xmark         & \cmark           & \xmark            & 0  & 0   & 0    & 0     & 0     & 14336    \\
\xmark         & \xmark            & \cmark           & 0  & 0   & 0    & 0     & 0     & 0        \\
\xmark         & \xmark            & \xmark            & 0  & 0   & 0    & 0     & 1120  & 0    \\ \midrule
\multicolumn{3}{c}{Total no. of functions} & 32 & 512 & 3840 & 17920 & 28000 & 14336  \\ \bottomrule
\end{tabular}
\end{table}

\begin{table*}[t]
\tiny
\centering
\caption{Percentage of functions with starting $nl_f$ where selected mutation operators can increase nonlinearity, $n=5$}
\label{tab:succ_n5}
\begin{tabular}{@{}ccccccccccccccc@{}}
\toprule
\multicolumn{3}{c}{Operator}               & \multicolumn{12}{c}{starting $nl_f$}            \\ \cmidrule(lr){1-3} \cmidrule(lr){4-15}
rot        & bit flip       & 2 bit flip      & 0  & 1   & 2    & 3     & 4     & 5 & 6 & 7 & 8 & 9 & 10 & 11       \\ \midrule
\cmark & \cmark & \cmark & 100 & 74 & 74 & 74 & 73 & 73 & 73 & 73 & 71    & 51    & 6     & 0   \\
\cmark & \cmark & \xmark  & 0   & 0  & 0  & 0  & 0  & 0  & 0  & 0  & 0     & 0.53  & 0     & 0   \\
\cmark & \xmark  & \cmark & 0   & 0  & 0  & 0  & 0  & 0  & 0  & 0  & 0     & 0     & 0     & 0   \\
\cmark & \xmark  & \xmark  & 0   & 0  & 0  & 0  & 0  & 0  & 0  & 0  & 0.001 & 0     & 0.097 & 0   \\
\xmark  & \cmark & \cmark & 0   & 26 & 26 & 26 & 27 & 27 & 27 & 27 & 29    & 47    & 90    & 0   \\
\xmark  & \cmark & \xmark  & 0   & 0  & 0  & 0  & 0  & 0  & 0  & 0  & 0     & 1.704 & 0     & 100 \\
\xmark  & \xmark  & \cmark & 0   & 0  & 0  & 0  & 0  & 0  & 0  & 0  & 0     & 0     & 0     & 0   \\
\xmark  & \xmark  & \xmark  & 0   & 0  & 0  & 0  & 0  & 0  & 0  & 0  & 0.003 & 0     & 3.62  & 0  \\ \bottomrule
\end{tabular}
\end{table*}

The same is evident for a higher number of variables; we present the results for $n=5$ in Table~\ref{tab:succ_n5}.
Here, the functions are only sampled (1\% of all functions), and the results are presented as percentages.
The last two tables provide an interesting observation: there are no functions for which \textit{only} two bit flip operator makes a transition to a higher nonlinearity.
For single bit flip and rotation, there are functions where only these mutations can improve nonlinearity.
This may indicate that the two bit flip operator is redundant in a local search algorithm, which is encouraging since its neighborhood size is considerably larger.
Still, the experiments do not always support this observation, as presented in Section~\ref{sec:exp_results}.

\subsection{Crossover Operators}

As the last experiment in operator analysis, we tested the effect of two crossover operators:
\begin{itemize}
    \item single-point crossover: break point always in the middle;
    \item uniform crossover: even bits from the first and odd from the second parent.
\end{itemize}

These operators were intentionally simplified to reduce the search size since we aim to investigate the influence that the nonlinearity of parent solutions has on the product of crossover.
We iterated over every pair of functions (or over a sample) and recorded the result of the crossover.
We differentiate three outcomes: the child solution has a greater $nl_f$ than both parents, lower than both, or some other value in between.
Naturally, we are mostly interested in the first outcome, so we only present the probabilities of the child having a greater nonlinearity.
We perform only sampling due to a large number of possible pairs, and the results for $n=4$  and both crossovers are presented in Tables~\ref{tbl:singlepoint} and~\ref{tbl:uniform}.
The row indexes correspond to the nonlinearity of the first parent and columns of the second one, while the values present probability in percentages.

\begin{table}
\small
\centering
\caption{Probability of increasing the nonlinearity with single-point crossover}
\label{tbl:singlepoint}
\begin{tabular}{@{}lllllll@{}}
\toprule
$nl_f$  & 0  & 1  & 2  & 3  & 4  & 5 \\ \midrule
0 & 71 & 74 & 66 & 21 & 0  & 0 \\
1 & 91 & 93 & 87 & 51 & 13 & 0 \\
2 & 78 & 84 & 87 & 57 & 19 & 1 \\
3 & 27 & 46 & 57 & 62 & 22 & 1 \\
4 & 0  & 11 & 18 & 22 & 24 & 1 \\
5 & 0  & 0  & 0  & 1  & 1  & 1 \\ \bottomrule
\end{tabular}
\end{table}

\begin{table}
\centering
\small
\caption{Probability of increasing the nonlinearity with uniform crossover}
\label{tbl:uniform}
\begin{tabular}{@{}lllllll@{}}
\toprule
 $nl_f$ & 0  & 1  & 2  & 3  & 4  & 5 \\ \midrule
0 & 85 & 85 & 76 & 27 & 0  & 0 \\
1 & 88 & 91 & 85 & 46 & 11 & 0 \\
2 & 80 & 84 & 87 & 56 & 18 & 0 \\
3 & 31 & 47 & 57 & 63 & 22 & 1 \\
4 & 0  & 10 & 18 & 22 & 24 & 1 \\
5 & 0  & 0  & 0  & 1  & 1  & 1 \\ \bottomrule
\end{tabular}
\end{table}

It is interesting to note that the probabilities of ``success'' of both operators are very similar.
This indicates that obtaining a higher $nl_f$ depends solely on the nonlinearity of the parents, at least for these two operators.
Additionally, highly successful crossovers occur only for parents of lower nonlinearity and when both parents have a similar nonlinearity.
From the algorithm perspective, this would indicate it is beneficial to perform crossover in the population of low-quality individuals, but for higher nonlinearity levels, mutation operators would present a better choice.

\section{Search Algorithms}
\label{sec:search}

\subsection{Genetic Algorithm}

To evaluate the influence of operators and their combinations, we use a genetic algorithm as a baseline.
We employ a canonical genetic algorithm with bit-string encoding and a steady-state replacement strategy. In each iteration, three solutions are selected at random and placed into a tournament.
The worst solution from the tournament is removed from the population; the remaining two individuals are used as parents to create a replacement solution.
The child solution then undergoes a mutation with a given probability and is added to the population.
The algorithm's parameters are as follows: the population size is 100, and the crossover is performed either as a uniform crossover (with every gene inherited from both parents with equal probability) or a single-point crossover (with any possible crossing point).
The mutation operators are a single bit flip mutation, two bit flip, and a mixing mutation, which randomly shuffles bits between two randomly chosen points; mutation is applied with the probability of 0.5 to every child solution. The parameters were selected based on the related works.

\subsection{Greedy Local Search}

Since, in the previous section, it was evident that only mutation operators can reach high nonlinearity solutions with relatively large probabilities, we employ the operators in a greedy local search (LS) algorithm.
The first LS variant (Algorithm~\ref{alg:ls}) iterates over the selected mutation operators (corresponding to different neighborhoods) and accepts the \textit{first solution} with a better fitness function - thus being greedy. If a better solution is found, the search is iterated with the new solution until no improvement is possible.
This algorithm is applied with different combinations of operators, as well as different \textit{orders} of operators within the main loop since that may also affect the outcome.

\begin{algorithm}[t]
\footnotesize
\caption{A greedy local search\label{alg:ls}}
\begin{algorithmic}[1]
  \REPEAT
    \FOR{each mutation operator}
        \FOR{each mutation position}
        \IF{better solution found}
          \STATE accept new solution and continue the main loop
        \ENDIF
        \ENDFOR
    \ENDFOR
  \UNTIL{there is no improvement}
\end{algorithmic}
\end{algorithm}

The last section showed that there are Boolean functions for which even the combined neighborhood of all three operators does not contain any Boolean function with a higher nonlinearity.
Therefore, we modify the simple LS: if there is no improvement in the current solution, the algorithm \textit{reverts} to the previous solution and \textit{continues} exploring its neighborhood.
In other words, when the algorithm reverts, it does not repeat the move that resulted in a dead-end but continues iterating over unexplored mutation positions and subsequent operators.
This modified version of the algorithm is denoted as LS-revert (LS-R) in the experiments.

\section{Experimental Results}
\label{sec:exp_results}

All the algorithms are run for a maximum of 500000 evaluations (or until convergence for the LS), and each experiment is repeated for 30 runs.
Additionally, the GA and LS variants are used with both fitness functions.
The experiments are conducted on Boolean sizes of 8, 9, and 10 variables; these are the sizes commonly used in related works and the smallest ones for which the optimal nonlinearity values may not be known.\footnote{For instance, for Boolean functions with 8 inputs, it is postulated that the maximal nonlinearity equals 118 when the function is balanced. Yet, the best-known result is 116.}

\subsection{Local Search Operators}

\begin{figure}
    \centering
    \includegraphics[trim={2.95cm 2.95cm 0 0},clip,width=1.0\linewidth]{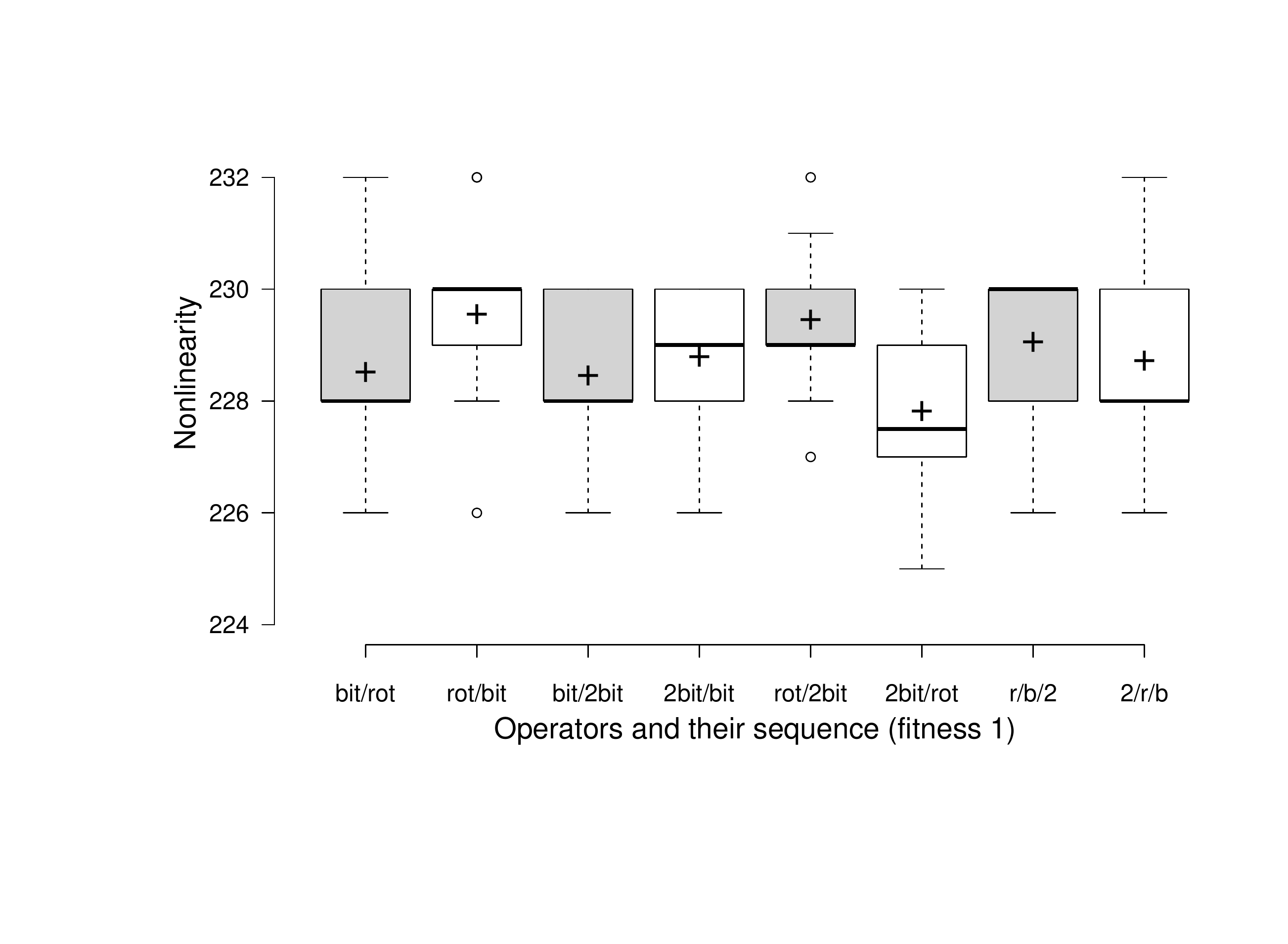}
    \caption{Performance of greedy LS with fitness 1 and different operator combinations (``bit'': bit flip, ``2bit'': two bit flip, ``rot'': rotation)}
    \label{fig:mut_fit1}
\end{figure}

First, we experimented with different combinations of mutation operators (neighborhoods) in the simple LS algorithm.
As mentioned before, the analysis indicated that only bit flip and rotation should be sufficient, while the two bit flip may be redundant.
As an example, we test various operator combinations and their sequences, using the first fitness function (Eq.~\eqref{eq:first}) on Boolean functions with 9 variables; the results of this experiment are shown in Fig.~\ref{fig:mut_fit1}.

\begin{figure}
    \centering
    \includegraphics[trim={2.95cm 2.95cm 0 0},clip,width=1.0\linewidth]{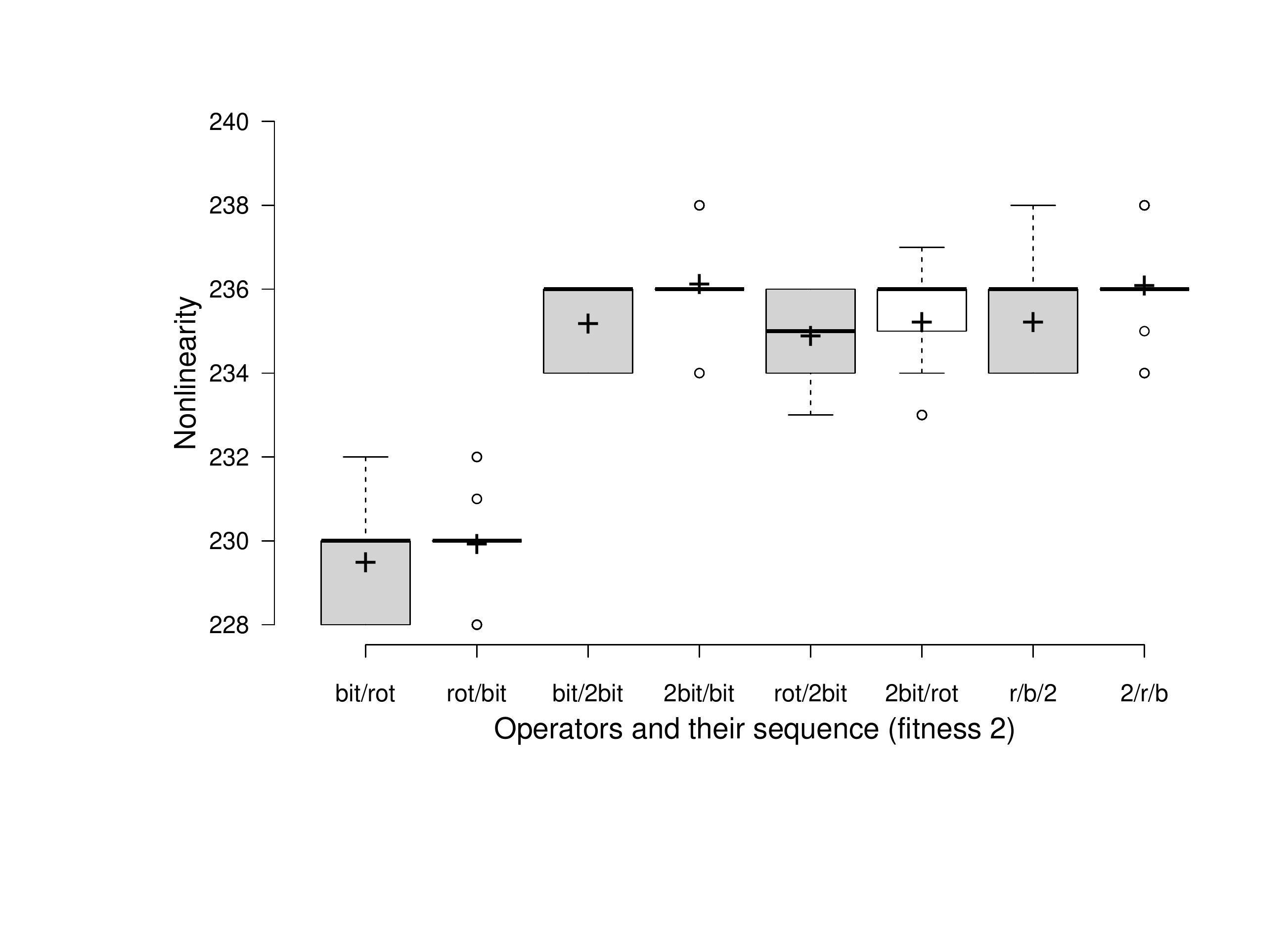}
    \caption{Performance of greedy LS with fitness 2 and different operator combinations (``bit'': bit flip, ``2bit'': two bit flip, ``rot'': rotation)}
    \label{fig:mut_fit2}
\end{figure}

Notice there are no significant differences between different operator combinations; also, the two bit flip seems not to affect the performance. 
However, the situation changes substantially when we use the same algorithm with the second fitness; the results for this case are presented in Fig.~\ref{fig:mut_fit2}.
It is evident that the second fitness function performs much better (see the scale on both figures).
Also, the inclusion of two bit flip operator induces a large improvement over the other variants, which is in contrast with the analysis that relied solely on nonlinearity value.
Additionally, the variants with two bit flip as the first used operator yield slightly better results.
Based on these results, any combinations with two bit flip could be chosen as the best performing one, and the combinations of ``2bit/bit'' and ``2bit/rot/bit'' yield the best mean value.
For algorithm comparison, we selected the variant with two bit flip and single bit flip since it includes only two operators.
While it may seem counter-intuitive to use both a single and two bit flip, note that multiple bit flips are performed at once, so it is possible that one operator succeeds where the other fails.

\begin{figure}
    \centering
    \includegraphics[trim={2.95cm 2.95cm 0 0},clip,width=1.0\linewidth]{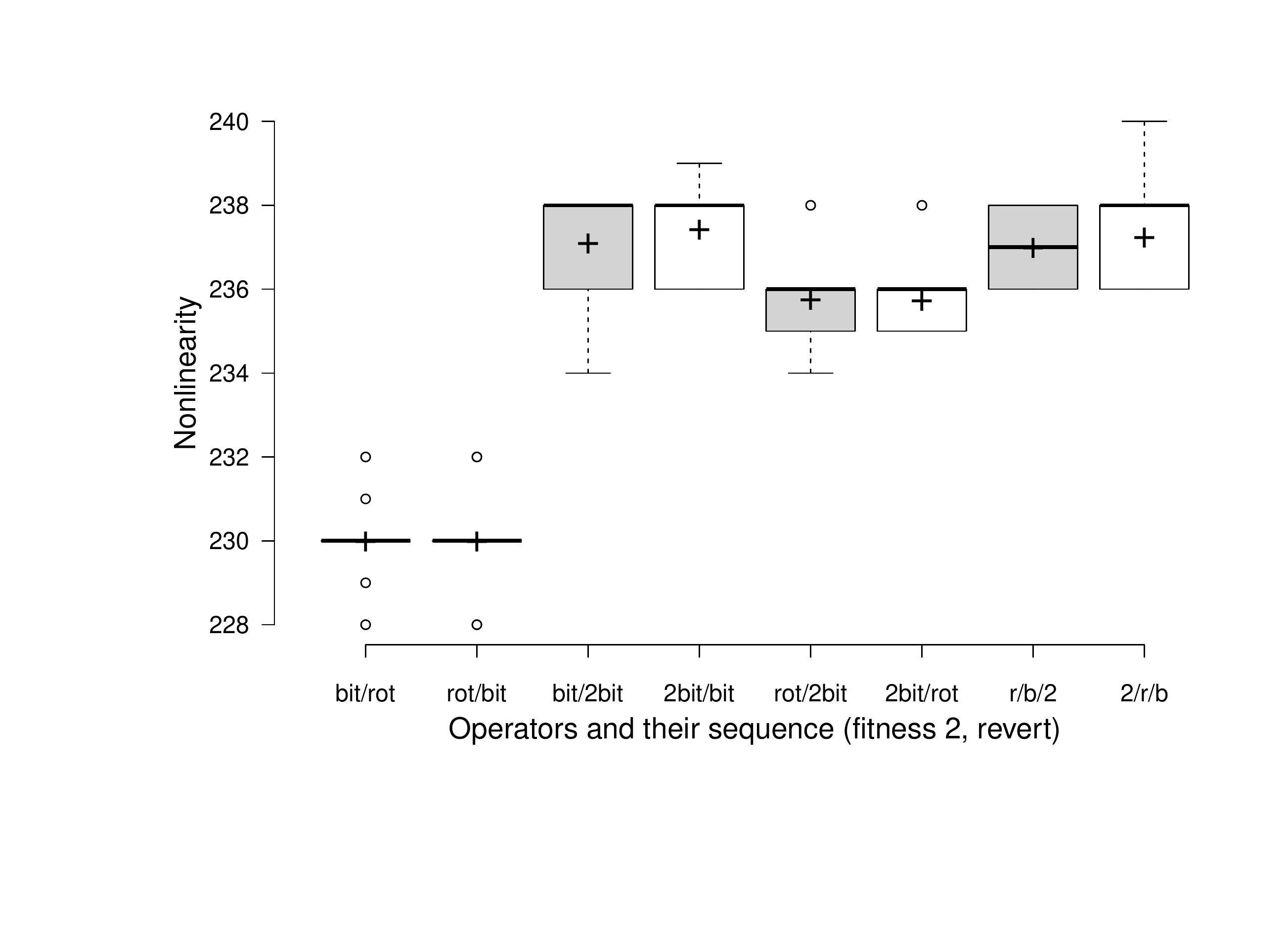}
    \caption{Performance of greedy LS with revert, fitness 2 and different operator combinations (``bit'': bit flip, ``2bit'': two bit flip, ``rot'': rotation)}
    \label{fig:mut_fit2_rev}
\end{figure}

Finally, the same operator combinations are applied with the LS-revert algorithm, for which the results are shown in Fig.~\ref{fig:mut_fit2_rev}.
This algorithm variant obtains better solutions for every operator combination (the nonlinearity scale is the same in both figures). 
This is expected and obtained at the cost of additional evaluations but still under the maximum evaluation limit.

\subsection{Algorithm Comparison}

\begin{figure}
    \centering
    \includegraphics[trim={2.95cm 2.95cm 0 0},clip,width=1.0\linewidth]{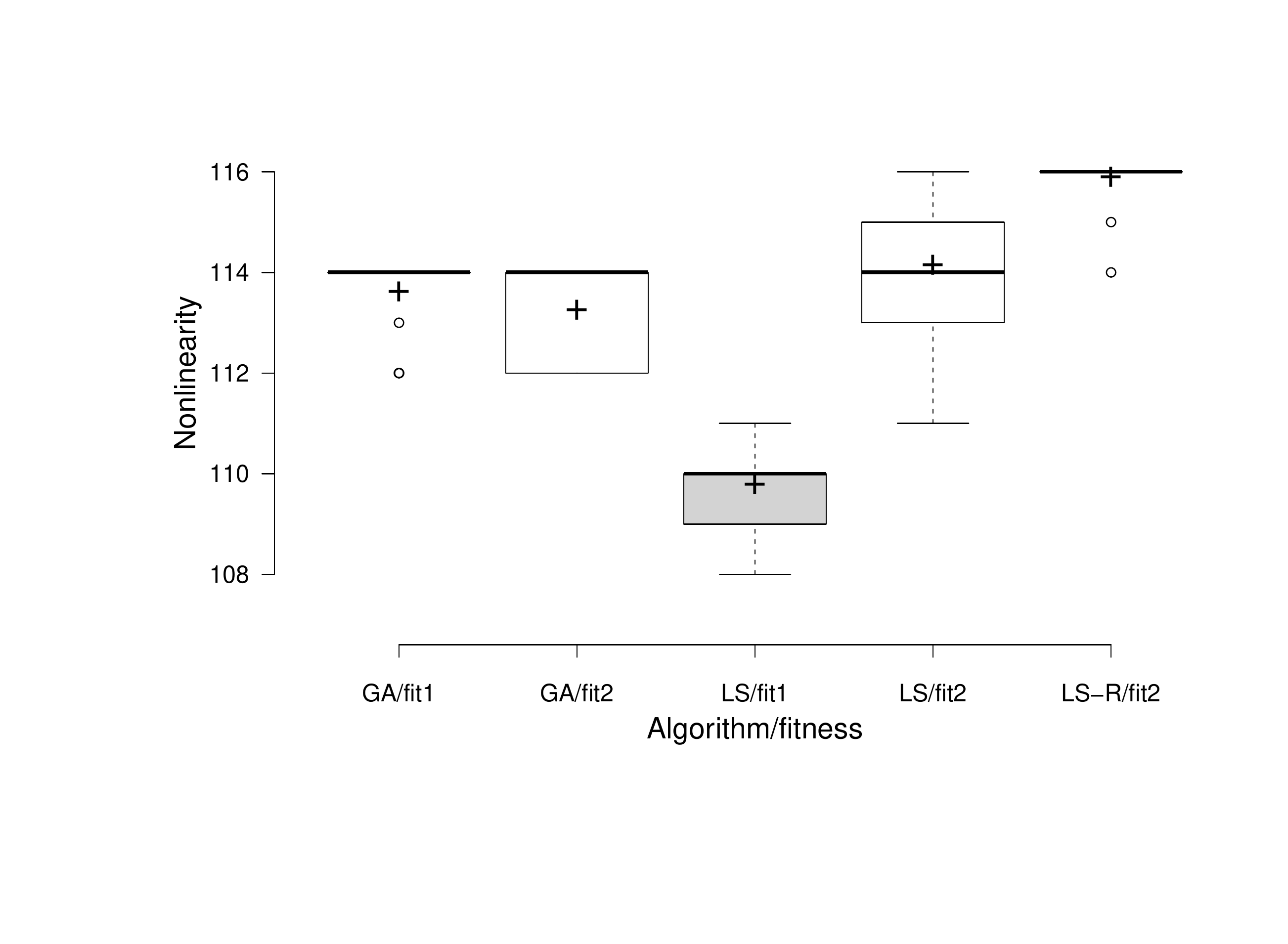}
    \caption{Algorithm comparison, $n = 8$}
    \label{fig:n8}
\end{figure}

\begin{figure}
    \centering
    \includegraphics[trim={2.95cm 2.95cm 0 0},clip,width=1.0\linewidth]{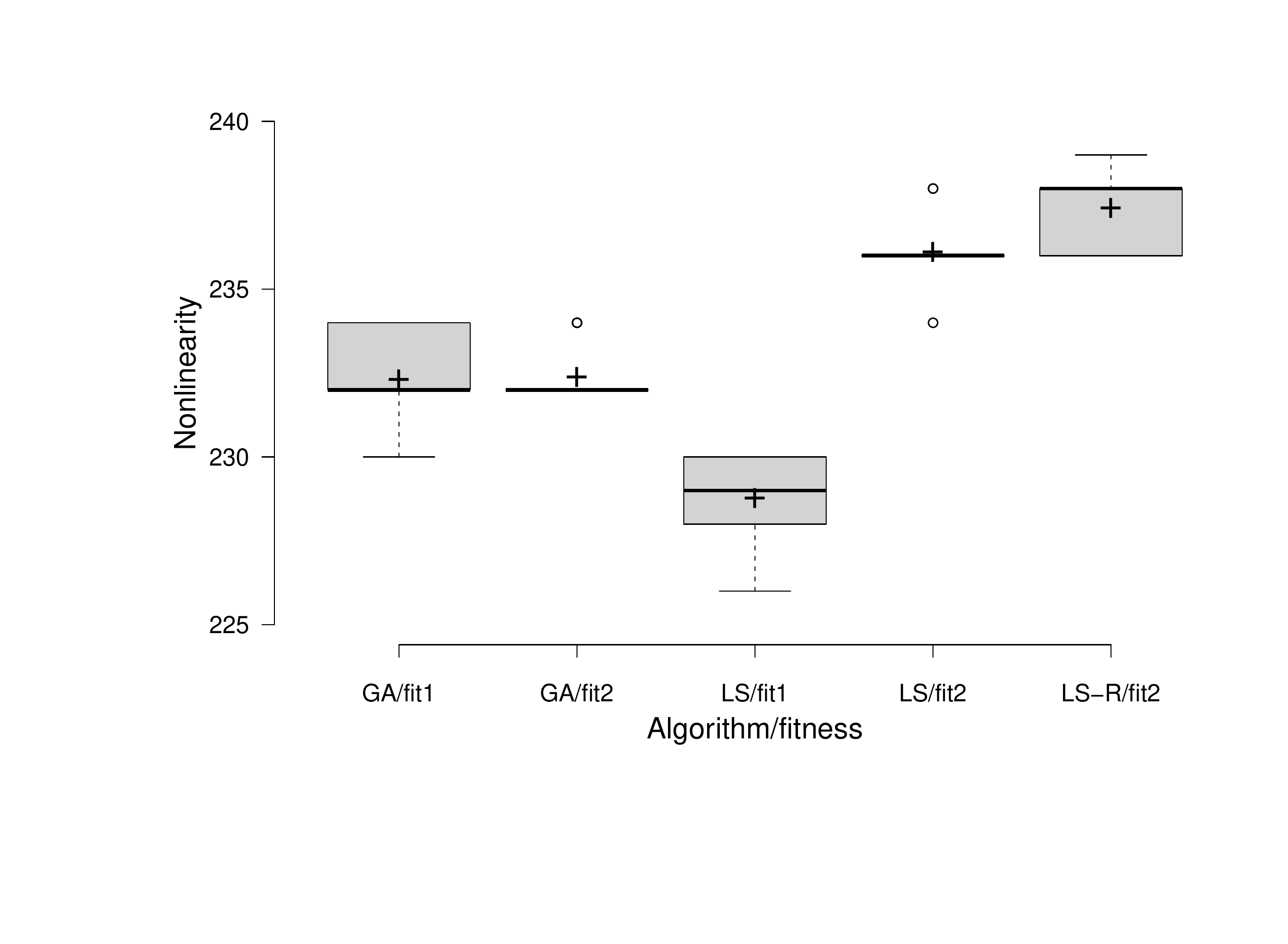}
    \caption{Algorithm comparison, $n = 9$}
    \label{fig:n9}
\end{figure}

\begin{figure}
    \centering
    \includegraphics[trim={2.95cm 2.95cm 0 0},clip,width=1.0\linewidth]{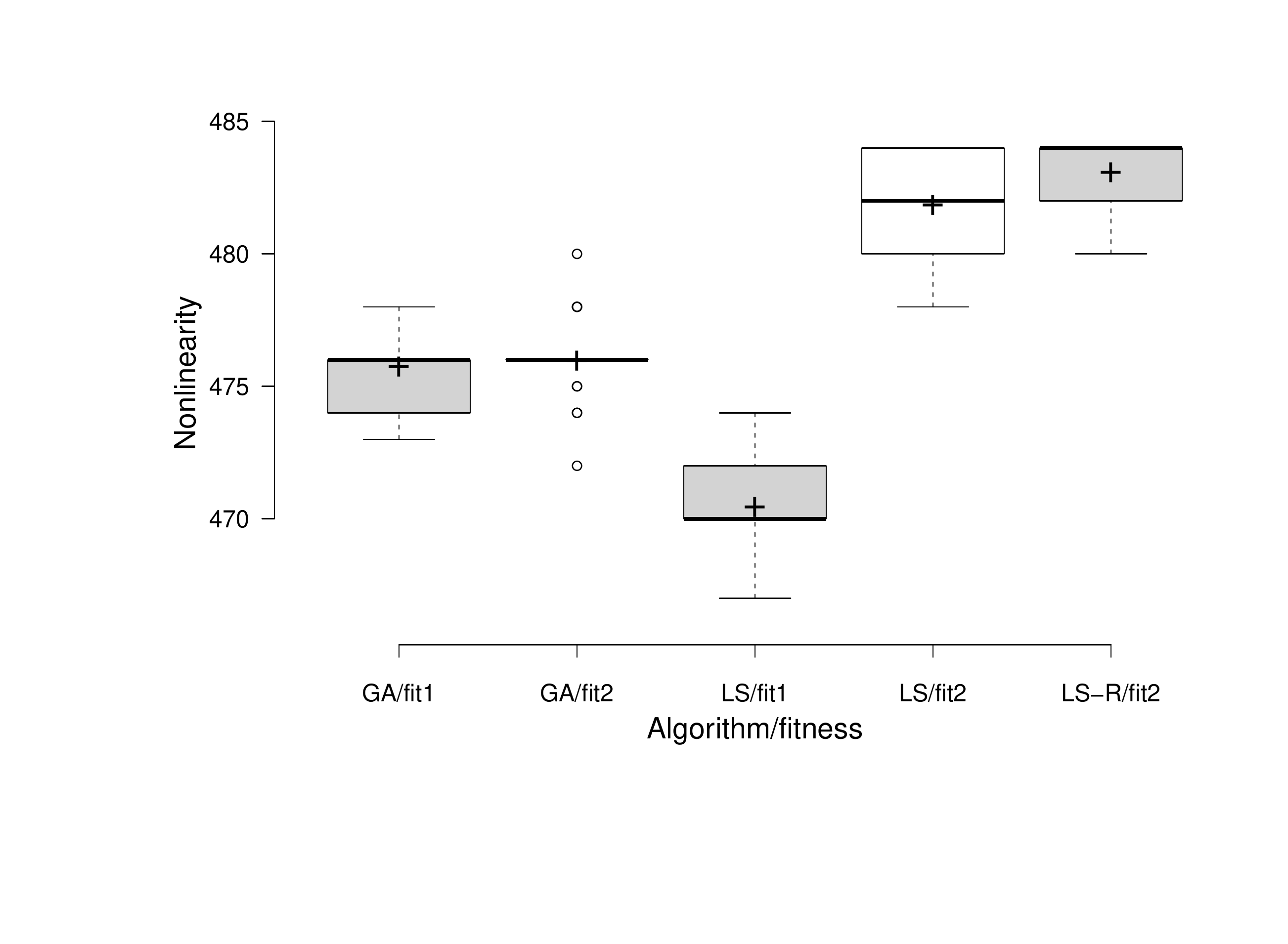}
    \caption{Algorithm comparison, $n = 10$}
    \label{fig:n10}
\end{figure}

In the second phase, the GA and LS variants are compared on different function sizes and using both fitness functions.
The results for all the algorithms in $n=8, 9$, and $10$ variables are shown in Figs.~\ref{fig:n8},~\ref{fig:n9}, and~\ref{fig:n10}.
Unlike the local search, GA does not benefit from using the second fitness function, at least not for smaller Boolean sizes.
Although the GA obtains worse results, this could be improved with careful parameter tuning, which was not performed in this analysis.
Still, it is clear that even a simple greedy local search offers competitive performance while using no other parameters besides the chosen operator set (which is, in this case, two bit flip and bit flip).
An evident downside of the LS approach is the prohibitively increasing time complexity with the increase of the solution size, but this can always be controlled (as in our experiments) with a set time or evaluation limit.

Based on the obtained results, we make the following general observations:
\begin{compactitem}
 \item Several simple mutation operators result in a consistent change in spectrum values for any Boolean function.
\item Not all mutation operators can make the transition to a Boolean function with a higher nonlinearity.
\item The results for bit flip and two bit flip show that over all functions with the same initial nonlinearity, the cumulative probabilities of nonlinearity changes are the same regardless of the mutation position, while the probabilities for the rotation operator depend on the number of rotated bits.
\item Using simple mutation operators, it is not necessarily possible to reach higher nonlinearity, but this can be resolved by using a more informative fitness function.
\item Obtaining a higher $nl_f$ depends on the parents' nonlinearity and not on the choice of a simple crossover operator. Highly successful crossovers occur only for parents of lower nonlinearity and when both parents have a similar nonlinearity.
\item LS benefits more than GA from a more informative fitness function when the Boolean function size is relatively small.
\end{compactitem}

The above observations might be related to the structure of the space of Boolean functions, especially when considering random sampling. Olejar and Stanek proved that the cryptographic properties of random Boolean functions are relatively close to the optimal values~\cite{OlejarS98}. This also holds for nonlinearity, and it might explain why the operators analyzed in this work cannot always improve the nonlinearity of the candidate solutions: already from the initial generation, they could be very close to a local optimum. 

The behavior of crossover can be explained by the geometry of the space of Boolean functions. In particular, we considered \emph{geometric operators}, meaning that they produce offspring lying on the segment between the two parents.
Applying crossover to these individuals could have a greater chance of sampling something on the border of non-overlapping spheres defined by two linear functions and thus obtain offspring with high nonlinearity.

\section{Conclusions and Future Work}
\label{sec:conclusions}

This paper investigates the effect of commonly used mutation and crossover operators on the Boolean functions' bit-string representation when optimizing nonlinearity.
Our findings show that the effects of different operators depend mainly on the starting nonlinearity of the Boolean function being affected rather than on the intrinsic operator parameters. The investigated crossover operators behave almost identically and offer no comparative advantage over mutation operators.

In future work, it would be interesting to investigate symbolic encoding as it provides better results than bit-string. Thus, we would consider the effect of changes there, from the tree encoding, into the truth table and, finally, the Walsh-Hadamard spectrum. Finally, a natural extension would be to consider S-boxes (vectorial Boolean functions), as there, one needs to assess all non-trivial combinations of Boolean functions, making the mapping between genotype and phenotype even more complex.

\bibliographystyle{abbrv}
\bibliography{bibliography}

\end{document}